\documentclass[conference]{IEEEtran}
\IEEEoverridecommandlockouts
\usepackage{cite}
\usepackage{amsmath,amssymb,amsfonts}
\usepackage{algorithmic}
\usepackage{graphicx}
\usepackage{textcomp}
\usepackage{xcolor}
\usepackage{multirow}
\usepackage[table,xcdraw]{xcolor}
\usepackage{enumitem}
\usepackage{graphicx}
\usepackage{subcaption}
\usepackage{booktabs}
\usepackage{comment}
\usepackage{pgfplots}
\usepackage{titlesec} 
\pgfplotsset{compat=1.18}
\usepackage[autostyle]{csquotes}
\MakeOuterQuote{"}
\usepackage{multirow}
\usepackage{graphicx}
\usepackage[table]{xcolor}
\definecolor{baselineRed}{RGB}{248,105,107}
\definecolor{fineGreen}{RGB}{99,190,123}
\definecolor{fineLightGreen}{RGB}{169,208,142}

\usepackage{xcolor}
\usepackage{pgfplots}
\pgfplotsset{compat=1.18}

\definecolor{shapeone}{RGB}{0,70,255}
\definecolor{shapetwo}{RGB}{255,127,14}
\definecolor{shapethree}{RGB}{0,150,60}
\definecolor{shapefour}{RGB}{230,30,30}
\definecolor{shapefive}{RGB}{190,0,90}

\def\BibTeX{{\rm B\kern-.05em{\sc i\kern-.025em b}\kern-.08em
    T\kern-.1667em\lower.7ex\hbox{E}\kern-.125emX}}
\begin{document}

\title{Fine-Tuning VLM for Enhancing AI's Spatial Intelligence: Understanding 3D and 2D Rotations \\

}

\author{\IEEEauthorblockN{Uttamasha Monjoree}
\IEEEauthorblockA{\textit{Department of Architecture} \\
\textit{Texas A\&M University}\\
College Station, USA \\
uxm190002@tamu.edu}
\and

\IEEEauthorblockN{Wei Yan}
\IEEEauthorblockA{\textit{Department of Architecture} \\
\textit{Texas A\&M University}\\
College Station, USA \\
wyan@tamu.edu}
}
\maketitle

\begin{abstract}
Spatial intelligence is a fundamental skill in multiple domains, such as Science, Technology, Engineering, and Mathematics (STEM), Medicine, Architecture, Construction, and Manufacturing. Recent studies indicate that Vision-Language Models (VLMs) still face limitations in spatial reasoning, which inhibits artificial intelligence (AI) from performing practical spatial tasks. Using multiple object-rotation datasets developed for training and evaluation, our experiments demonstrated promising improvements in both 2D and 3D rotation detection. Fine-tuned Google DeepMind-built Gemma-4 mixture-of-experts (MoE) models significantly outperformed fine-tuned Gemma-4 generalist models in predicting rotations defined by both their axes and angles. Fine-tuning also substantially improved angle estimation for 2D representation without requiring an explicit coordinate system. Furthermore, identifiable objects did not improve angle-detection accuracy; instead, objects with prominent linear features showed improved performance.
\end{abstract}


\section{Introduction}

Spatial abilities play a critical role in education and professional practice in many fields, such as AEC, STEM, and medicine. They are also essential for tasks such as assembly, construction, robotics, and medicine, where accurate interpretation and manipulation of three-dimensional information are required. A key component of spatial ability is spatial visualization, which is the capacity to mentally rotate, transform, twist, or invert 3D objects \cite{ben1988effect}, \cite{fleisig2004improving}. As a result, spatial visualization, as part of spatial skill, has important implications for both learning and real-world task performance.

Generative Artificial Intelligence (AI) technology has rapidly evolved in recent years, driven by advancements in deep learning and neural network architecture.
It can create new content such as text, images, 3D models, and audio. Large language models (LLMs) such as GPT-5 and its successors have demonstrated remarkable proficiency in generating coherent and contextually relevant text. GPT-5 with vision has the capability to analyze images and shows an unprecedented ability to understand visual inputs and the spatial relationships of humans and objects \cite{achiam2023gpt},\cite{han2026decoupling}.

\par
Many STEM students continue to experience difficulty with spatial visualization tasks even though many computer-assisted learning tools for geometry, spatial transformations, and related mathematical concepts are available to them \cite{mystakidis2022systematic}, \cite{garmendia2007first}.
Emerging technologies, including augmented reality, virtual laboratories, and AI-driven assistance, enable students to explore STEM concepts at their own pace while developing critical problem-solving skills through interactive and engaging learning experiences \cite{torres2026impact, al2023analyzing}. With conversational AI integrated into an immersive visual learning approach, learners can receive personalized, real-time assistance, allowing them to communicate, correct misunderstandings, and practice spatial reasoning interactively \cite{monjoree_ar-classroom_2024}.

\par
AI vision models can play an important role in disaster management, real-time incident detection, healthcare emergency response, industrial hazard prevention, and intelligent traffic control \cite{bajwa2025ai}. AI vision models need to be precise in spatial skills, as there is no room for error in these critical tasks. However, current vision-language models (VLMs) still have limitations in spatial reasoning, especially when interpreting three-dimensional (3D) transformations such as rotation, translation, scale, orientation, and perspective changes \cite{han2020spare3d}. AI systems can describe images and objects, but they are not yet consistently reliable as spatial guides in complex real-world environments where decisions may affect safety, infrastructure, and human life \cite{anthi2025role}, \cite{lee2025agentic}. 
Advancing the spatial intelligence of AI is critical to developing reliable human-AI interaction systems that are capable of assisting users in hazardous, time-sensitive, and visually complex environments \cite{speiser2026agentic}.

\par
Integrated verbal and visual descriptions can be used to explain spatial rotations and convey how objects move or change orientations in 3D space \cite{lacey2013multisensory}. AI's understanding of spatial rotation can be improved by combining the visual and verbal abilities of conversational AI \cite{monjoree_ar-classroom_2024}, \cite{11050695}. Previous studies have shown that VLMs struggle with spatial reasoning tasks despite using methods like fine-tuning and Chain of Thought (CoT) \cite{xu-etal-2025-defining}. The aim of this study is to enhance AI’s capability of understanding 3D spatial rotations by training a currently available VLM through a fine-tuning method and performing a series of experiments with 3D and 2D views of object rotation images. The research questions that guide this study are as follows:

\begin{enumerate}
    \item What is an efficient process for improving generative AI models' performance for understanding rotation via fine-tuning?
    \item How does the performance of the fine-tuned AI model for understanding rotation compare with the pre-trained model?
    \item How does the performance of fine-tuned AI differ in detecting rotation based on identifiable objects and arbitrary objects?
    
\end{enumerate}

\section{Literature Review}

\subsection{Angle Discrimination Ability of Humans \& Dual-coding Theory} The angular just-noticeable difference (JND) represents the smallest change in orientation that a human observer can reliably discriminate. Under ideal conditions, the JND for human angle discrimination is roughly 2° to 5° \cite{xu2018human}. Under controlled viewing conditions involving high-contrast lines or gratings, orientation-discrimination thresholds can be as small as approximately 0.15° to 0.5°, demonstrating a form of visual hyperacuity \cite{regan1985postadaptation}. Humans generally discriminate changes around horizontal and vertical orientations more accurately than changes around oblique orientations, a phenomenon known as the oblique effect \cite{orban1984orientation}, \cite{heeley1996mechanisms}. AI has the potential to be trained for its ability to understand orientation differences and angle discrimination to guide us in spatial reasoning tasks based on the knowledge of human cognition. \par
The dual-coding theory describes the connection between two subsystems in the human cognitive system: an imagery system and a verbal system \cite{clark1991dual}. Human cognition integrates language and nonverbal objects and events and helps us to process linguistic input (i.e., writing and speech) with symbolic representations of nonverbal aspects of the world \cite{paivio1990mental}.  Thus, spatial information represented by integrating these two systems, vision and language, can help AI direct communication and provide a framework for understanding nonverbal phenomena. Integrating verbal information with images can also help us perform angle discrimination tasks better.

\subsection{VLMs' Spatial Abilities}

Through multimodal pre-training, VLMs align visual features with natural-language concepts, enabling them to develop semantic representations that capture objects’ categories, attributes, functions, and relationships with other objects. CLIP demonstrated that image–text alignment enables natural language to reference learned visual concepts and supports zero-shot object recognition \cite{radford2021clip}. Similarly, BLIP-2 showed that connecting a pretrained vision encoder with a large language model enables instruction-guided image interpretation, visual question answering, and image-to-text generation \cite{li2023blip2}. 

Spatial intelligence in VLMs has the potential to guide users in comprehending visual-spatial tasks and spatial relations and assist in gaining a sense of orientation of objects in space \cite{hawes2022effects}. Basic spatial ability tests demonstrated that VLMs mirror human hierarchies (strongest in 2D orientation, weakest in 3D rotation) \cite{xu-etal-2025-defining}. Research indicates that while generative AI models GPT-4, Gemini 1.5 Pro, and Llama 3.2, lack an understanding of spatial reasoning, they have the potential to understand the rotation process when language/contextual information is utilized to process spatial information \cite{11050695}.

\subsection{AI Vision Model Applications}

AI's vision and spatial understanding abilities have a wide range of applications in STEM and AEC, such as intelligent optimization, site layout, risk assessment, Structural health monitoring, incident reporting, and compliance evaluation \cite{saka2024gpt, li2025applications}, \cite{chen2025tailored}, \cite{zhang2026cognitively}. Experiments have proven the GPT vision model to be highly effective in detecting and explaining global and fine-grained semantic patterns in anomaly or risk detection \cite{yang2023dawn}, \cite{driessen2024putting}. Several recent studies have focused on fine-tuning LLMs for diverse downstream applications, including interactive decision-making \cite{li2022pre} and tasks performed within interactive environments \cite{xiang2023language}. However, current VLMs still demonstrate limited capabilities in comprehensively interpreting 3D scenes and reasoning about the interactions that may occur within them. To address these limitations, several studies have developed specialized datasets for evaluating and improving spatial understanding and reasoning \cite{liu2023visual, michel2023object}.

Research on ChatGPT-4 and ChatGPT-4o in solving mathematics across grades 4, 8, and 12 shows that while AI can support algebra instruction, teacher intervention is needed in areas where AI is less effective, such as spatial intelligence and higher-difficulty tasks \cite{wei2024evaluating}. GSM-Symbolic enables more controllable evaluations and reliable metrics for measuring the reasoning capabilities of LLMs; it has been observed that LLMs resemble advanced pattern matching more than true logical reasoning \cite{mirzadeh2024gsm}. LLM performance varies widely across different versions of the same question and a minor increase in difficulty. As a result, the mathematical performance of AI underperforms when compared to a graduate student’s level \cite{frieder2024mathematical}. However, in recent years, AI has demonstrated the potential to improve learning by enhancing interactivity, accessibility, and personalization. AI- and AR-based multi-sensory assistance helps undergraduate students better understand spatial transformations and their mathematical representations \cite{monjoree_ar-classroom_2024}.

\subsection{VLMs in Training and Practice}

VLMs can be utilized for construction process regulation and safety measures. Fine-tuned VLMs for construction hazard report generation can automatically produce structured and detailed hazard reports and provide an innovative solution for construction safety management, enhancing efficiency, accuracy, and automation in construction hazard identification \cite{chen2025tailored}. TOPVIEWRS, a dataset based on realistic and semantic top-view maps, was used to evaluate VLMs for spatial-reasoning tasks \cite{li-etal-2024-topviewrs}. It revealed a performance gap of more than 50\% relative to humans, with some models performing below random chance; although CoT reasoning improves accuracy by 5.82\% on average, overall top-view spatial reasoning remains limited. Prompt modularization and hybrid optimization of VLMs provided a scalable and practical computational pipeline for VLM-assisted urban building safety supervision \cite{zhang2026cognitively}.

\subsection{Fine-tuning AI}

Fine-tuning adapts a pre-trained model’s learned patterns and features to new tasks, enhancing performance and minimizing the need for extensive training data \cite{parthasarathy2024ultimate}. It has gained popularity in generative AI models for tasks such as text and image classification, object detection, sentiment analysis, and question-answering. Supervised fine-tuning (SFT) involves training LLMs on labeled data specific to the target task. For instance, fine-tuning an LLM for text classification in a business context would use a dataset of text snippets with corresponding class labels \cite{parthasarathy2024ultimate}. Although effective, this approach still demands a significant amount of labeled data, which can be both expensive and time-consuming to gather.

\section{Methodology}

In this paper, we focused on enhancing AI's spatial ability by fine-tuning a VLM based on prior work that evaluated pretrained AI models' spatial ability \cite{11050695}, \cite{monjoree2024aisspatialintelligenceevaluating}. 

 \par

 \par
We used Gemma-4 by Google for all our experiments in this paper. A previous study demonstrated that superimposing additional context in the object rotation dataset improved AI's understanding of 3D rotation \cite{11050695}. We adopted two methods to create our dataset images in this paper for fine-tuning the AI model's vision capability to identify (1) the 3D object rotation process based on visual cues, (2) the 2D object rotation process without visual cues.

\textbf{(1) 3D Rotation Process w/ Visual Cues:} This method is based on the dual-coding theory, where we used visual cues on 3D objects for fine-tuning and evaluation purposes. Dual-coding posits how imagery and verbal cognitive subsystems work together to process language and non-verbal content. While describing rotation, we use references like direction, angle, and axis visually and verbally. Integrating visual and verbal systems into the fine-tuning process has the potential to enhance spatial understanding, where language aids in communication and vision aids in interpreting nonverbal cues. We focused on identifying performance differences of VLM before and after fine-tuning based on two different fine-tuning methods. Experiment 1.1 \& Experiment 1.2 in the later sections are implemented using this method.

\textbf{(2) 2D Rotation Process w/o Visual Cues:} In this method, we used a planar view (top view) of 3D objects to create our dataset for fine-tuning and evaluation purposes. We focused on identifying performance differences of the VLM before and after fine-tuning, along with evaluating its performance based on identifiable and arbitrary objects. Experiments 2.1 \& 2.2 in the later section are implemented using this method.

\par
 \par

We used supervised fine-tuning (SFT), which involves training VLMs on labeled data specific to the target task. Here, we fine-tuned a VLM for image classification to identify spatial data with class labels (axis of rotation, angle of rotation) (Fig \ref{fig:ftflowchart}). In all experiments of this paper, we used Parameter-Efficient Fine-Tuning (PEFT) methods, which use adapter layers to enable partial fine-tuning of the model \cite{liu2022few}. This approach adds extra layers (adapters) to the pre-trained model, allowing for more efficient fine-tuning with fewer parameters, and addressing challenges related to computational efficiency, overfitting, and optimization \cite{xu2026parameter}. For fine-tuning large neural networks, techniques like LoRA (Low-Rank Adaptation) is used \cite{zhang2023machine}, \cite{dettmers2023qlora}.

\section{Experiments }

\textbf{Dataset Generation:} \par
\begin{figure*}[t]
    \centering
    \includegraphics[width=.8\textwidth]{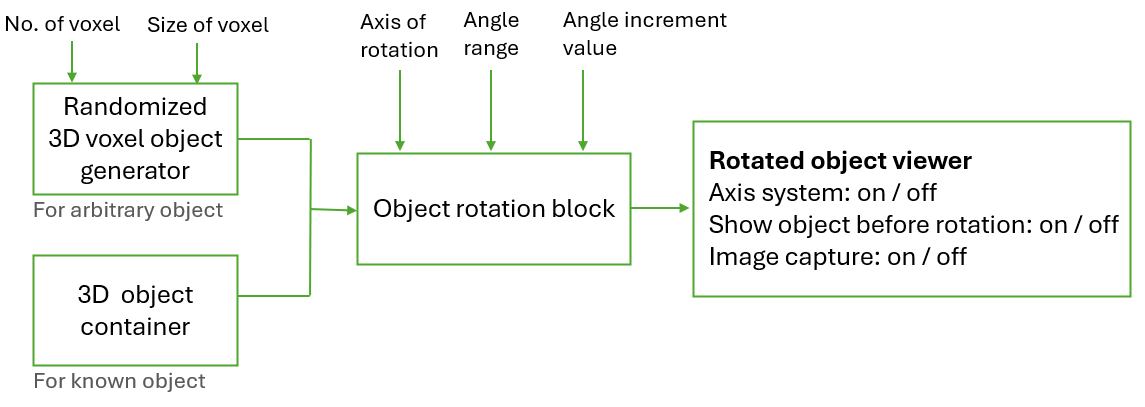}
    \caption{Workflow of rotation image generation in Grasshopper}
    \label{fig:3dflowchart}
\end{figure*}

We used 3D modeling software Rhino and its Grasshopper plugin for 3D object generation and rotation to create the dataset images, including isometric views and top views that are used in experiments. We implemented parametric modeling for arbitrary object generation using voxels, rotation implementation, and image capture (Fig. \ref{fig:3dflowchart}). We can customize the size, number of voxels, shape, and color for arbitrary object generation. After that, either the arbitrary object or any 3D object model can be attached to the object rotation block. The object rotation block takes the axis of rotation, angle range, and angle increment value as input. Then the rotated object viewer comes into effect. This viewer has options to show the axis system, the object before rotation, and image capture. When the object rotation block is turned on, it will rotate the object according to the set configuration and capture an image after each step.

\begin{figure}[htbp]
\centerline{\includegraphics[width=.8\linewidth]{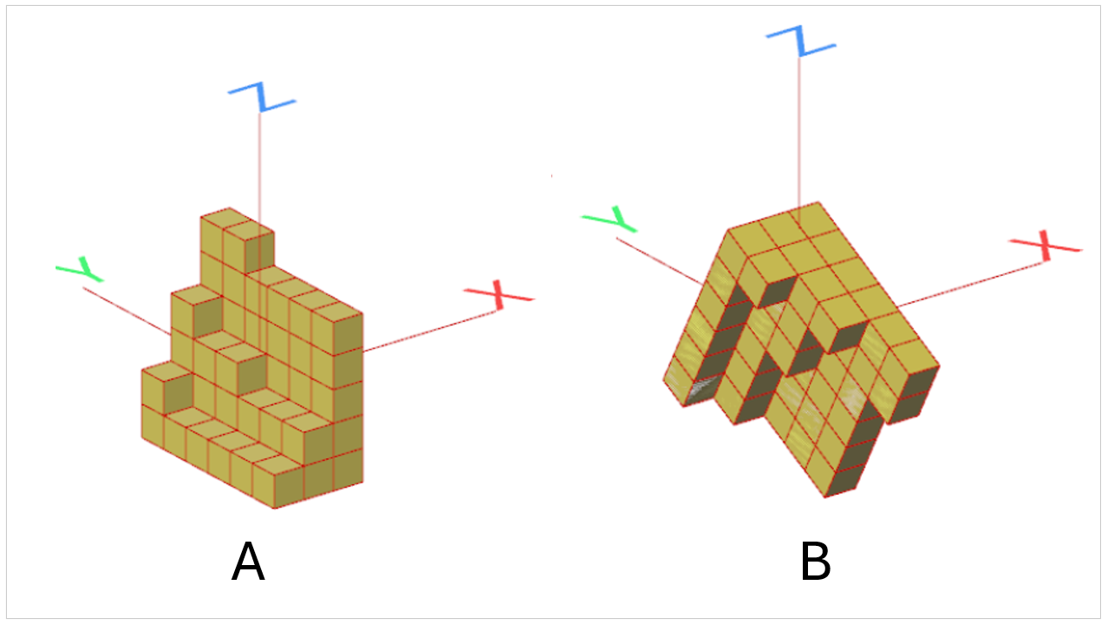}}
\caption{Example from 3D voxel dataset (used in Experiment 1.1 \& 1.2) where the object is rotating along the X-axis}
\label{fig:3Dvoxel}
\end{figure}

\textbf{(1) 3D Rotation Process w/ Visual Cues:}
LLMs rely on descriptive datasets and visual cues, which can enhance vision models' spatial visualization ability. We generated a dataset of 495 images of a 3D object (selected randomly from continuously incrementing angles from 0 to 180 degrees) rotating around three axes: X-, Y-, and Z-axis (Fig. \ref{fig:3Dvoxel}). Images were captured at varying distances from the camera, from slightly fluctuating viewpoints. The world coordinate system was superimposed on the 3D object (Fig. \ref{fig:3Dvoxel}). A (left-side) represents the object before rotation, and B (right-side) represents the object after rotation.

\textbf{\textit{Experiment 1.1 (Generalist Fine-tuning): Detecting Rotation Axis \& Angle Jointly: }} \par

We used the pre-trained model for the baseline test by randomly selecting 10\% as test data. Then we fine-tuned the base model with 90\% training data using the PEFT method mentioned above. We utilized the same test set for evaluation of the fine-tuned model.

System Prompt: \textit{``You are seeing two perspective views of an object. The left object is in a 0-degree rotation position. The right-side object is rotated clockwise. You are learning the axis and angle of rotation of the right-side object. For the axis, answer using X, Y, and Z. For the angle, only answer in increments of 1 degree, starting from 0 and ending in 180 degrees.''}
The query to the fine-tuned model is: \textit{ "A is the previous pose, and B is posed after rotation. Can you say which is the axis and angle of rotation?"}

\par

\textbf{\textit{Experiment 1.2 (MoE Fine-tuning): Detecting Rotation Axis \& Angle Separately:}} \par

As we did not achieve adequate accuracy with the generalist fine-tuned model (Table 1), we proceeded with the Mixture of Experts (MoE) method. It is a model architecture technique that speeds up AI inference by routing tasks to the most capable part of the model \cite{ibm_mixture_of_experts}. Here, we trained the AI for two tasks separately: axis detection and angle detection.

\textbf{\textit{Rotation Axis Detection: }} We fine-tuned the AI using PEFT with only axis data of three classes (X, Y, and Z).
System prompt: \textit{\enquote{You are seeing two perspective views of an object. The left object is at a 0-degree rotation. The right-side object is rotated clockwise. You are learning the axis of rotation of the right-side object. For the axis, answer using X, Y, and Z.}}

\textbf{\textit{Rotation Angle Detection:}} We fine-tuned the AI using PEFT with only angle data (180 classes).
System prompt: \textit{\enquote{You are seeing two perspective views of an object. The left object is at a 0-degree rotation. The right-side object is rotated clockwise. You are learning the angle of rotation of the right-side object. Only answer in increments of 1 degree, starting from 0 and ending in 180 degrees.}}

\textbf{(2) 2D Rotation Process w/o Visual Cues:}

The fine-tuned model for 3D images has satisfactory results in the case of axis detection (Table \ref{tab1}). However, angle detection still needs sufficient improvement. Therefore, we designed a simplified angle detection task.

In this experiment, we used planar images of 3D objects generated using the Dataset Generation method described previously, without any visual cues that help understand rotation. As we used a planar top view (X-Y plane), rotation only occurred around one axis (Z). So the coordinate system is not shown in this experiment. \par
\begin{figure}[t]
    \centering

    \begin{subfigure}{\columnwidth}
        \centering
        \includegraphics[width=\columnwidth]{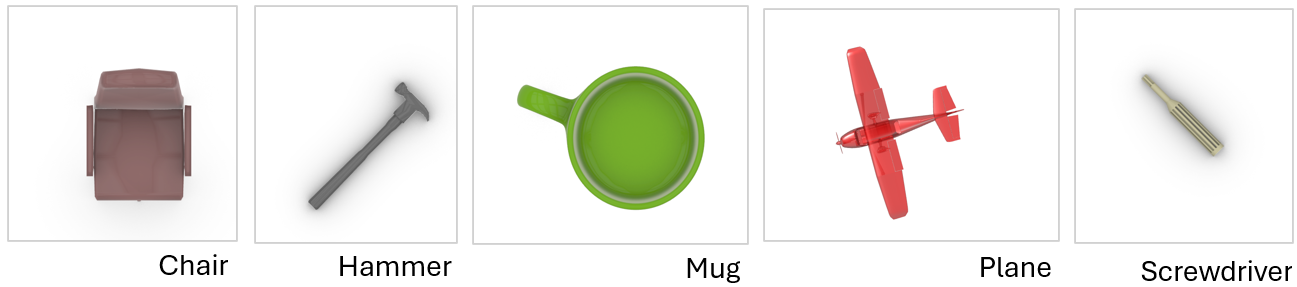}
        \caption{Object Examples}
        \label{fig:image_a}
    \end{subfigure}

    \vspace{3pt}

    \begin{subfigure}{\columnwidth}
        \centering
        \includegraphics[width=\columnwidth]{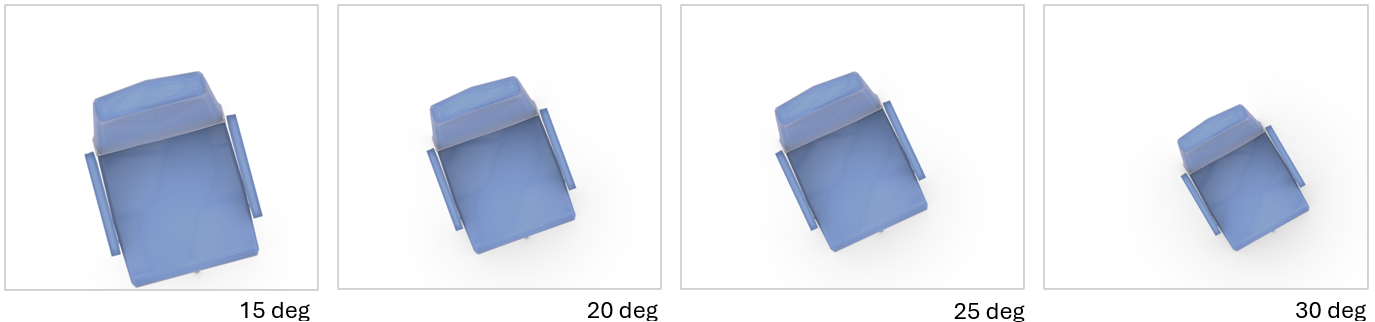}
        \caption{Examples from Training Dataset (Increment by 5 Deg)}
        \label{fig:image_c}
    \end{subfigure}

    \caption{Image Examples from Dataset of Identifiable Objects Used in Experiment 2.1.}
    \label{fig:known obj}
\end{figure}
\begin{figure}[t]
    \centering

    \begin{subfigure}{\columnwidth}
        \centering
        \includegraphics[width=\columnwidth]{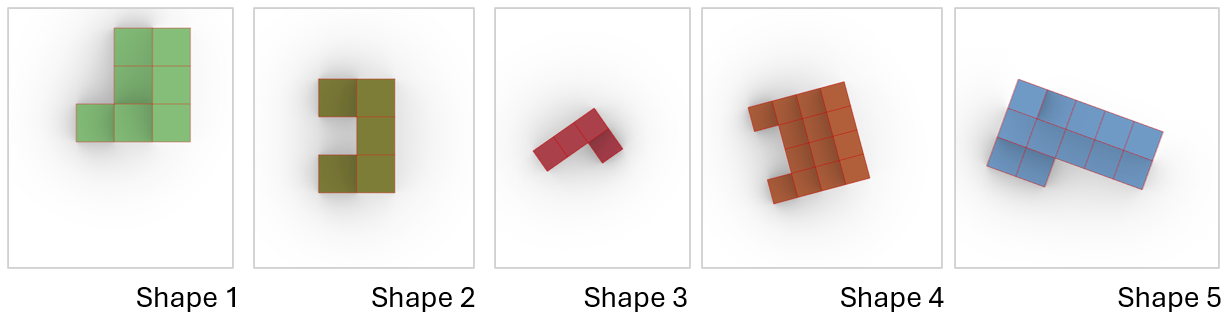}
        \caption{Object Examples}
        \label{fig:image_a}
    \end{subfigure}

    \vspace{3pt}

    \begin{subfigure}{\columnwidth}
        \centering
        \includegraphics[width=\columnwidth]{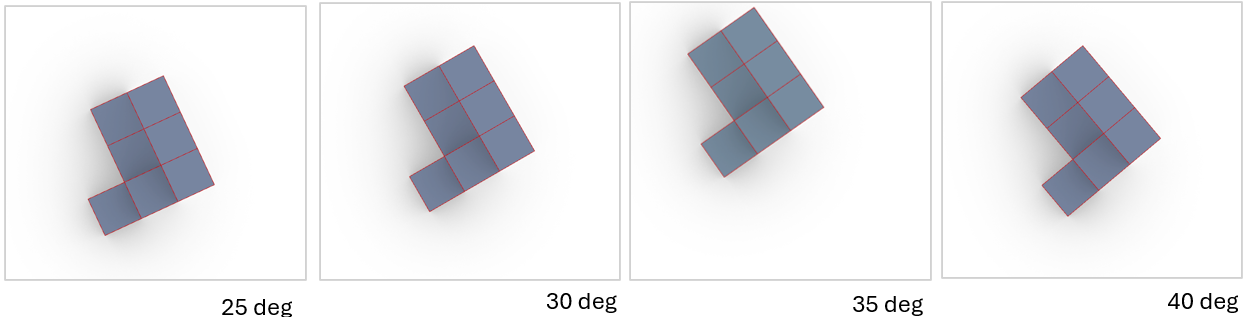}
        \caption{Examples from Training Dataset (Increment by 5 Deg)}
        \label{fig:image_c}
    \end{subfigure}

    \caption{Image Examples from Dataset of Arbitrary Objects Used in Experiment 2.2.}
    \label{fig:unknown obj}
\end{figure}

Rather than using only arbitrarily generated objects, we included everyday objects that AI can easily identify. Our two datasets include: (1) identifiable objects, e.g., chair, mug, etc. (Fig. \ref{fig:known obj}a); (2) arbitrarily generated 3D voxel objects (Fig. \ref{fig:unknown obj}a). The 3D models of the identifiable objects were adapted from models in multiple online sources \cite{unity3dassets}, \cite{sketchup3dwarehouse}. 

The prompt for the fine-tuned model was similar to the baseline test. We used a supervised PEFT method for fine-tuning the pre-trained model with 90\% of the remaining dataset. The same test set was used to evaluate both the pre-trained and fine-tuned models.
Query for this experiment was: \textit{"What is the angle of rotation?"} \par

\textbf{\textit{Experiment 2.1: Angle Detection along a Single Axis for Known Objects: }}

We used 5 object samples per dataset. Additionally, we decreased the number of rotation classes by incrementing the angles by 5 degrees, so we had 37 classes per object (Fig. \ref{fig:known obj}b). The datasets had 3000 images, where the angle was incremented by 5 degrees from 0 to 180 degrees. Each type of object, e.g., plane, varied in size, shape, and color. We did the baseline test using 10\% randomly selected data. The classes were $ 0^\circ $, $ 5^\circ $, $ 10^\circ $, \ldots, $ 180^\circ $.
System prompt example: \textit{"You are seeing the top view of an airplane. When the nose of the airplane is pointing left, and the body is completely horizontal, the angle of rotation is 0 degrees. The plane rotates anti-clockwise with an increment of 5 degrees. You are learning the angle of rotation. Only answer in increments of 5 degrees starting from 0 and ending in 180 degrees."} (An orientation convention would be the airplane nose pointing right being 0 degrees, but the test is focused on the changes of rotation angles.)

\textbf{\textit{Experiment 2.2: Angle Detection along a Single Axis for Arbitrary Objects: }}

We used 5 object samples per dataset. Additionally, we decreased the number of rotation classes by incrementing the angles by 5 degrees, so we had 37 classes per object (Fig. \ref{fig:unknown obj}b). The datasets had 3000 images, where the angle was incremented by 5 degrees from 0 to 180 degrees. Each type of object, e.g., shape 1, varied in size, shape, and color. We did the baseline test using 10\% randomly selected data. The classes were $ 0^\circ $, $ 5^\circ $, $ 10^\circ $, \ldots, $ 180^\circ $.
System prompt: \textit{"You are seeing the top view of an arbitrary object. The object rotates anti-clockwise with an increment of 5 degrees. You are learning the angle of rotation. Only answer in increments of 5 degrees starting from 0 and ending in 180 degrees."}

\begin{figure*}[t]
    \centering
    \includegraphics[width=.8\textwidth]{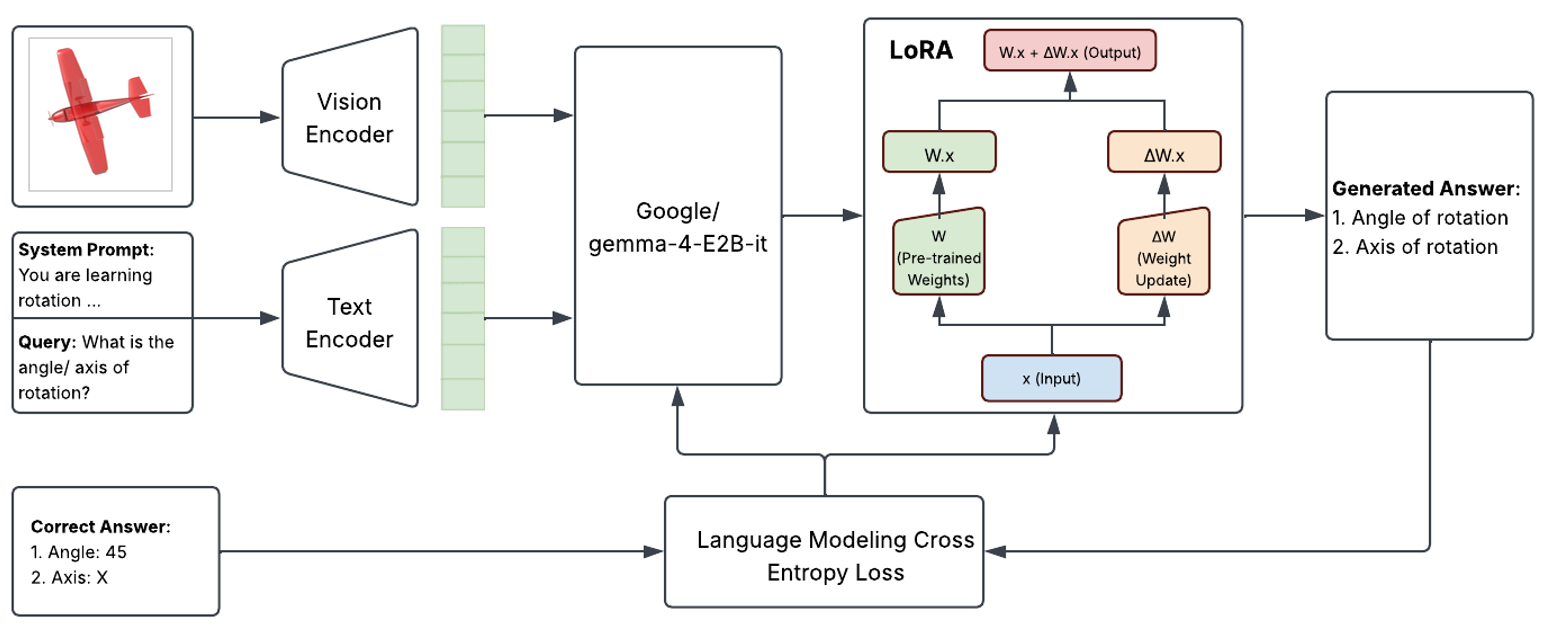}
    \caption{Multimodal image-text LoRA fine-tuning architecture for angle and axis of rotation prediction.}
    \label{fig:ftflowchart}
\end{figure*}
\par

\par
\textbf{Fine-tuning Procedure:} \par

For the experiments, our baseline model is Google/Gemma-4-E2B-it. We fine-tune it using PEFT (Fig. \ref{fig:ftflowchart}). Choosing appropriate evaluation metrics is essential to measure how closely the model’s predictions match the actual data and ensure the model is learning effectively \cite{parthasarathy2024ultimate}. The model was fine-tuned using LoRA as a parameter-efficient fine-tuning strategy. During fine-tuning, the vision encoder extracted image features while the language model encoded the accompanying textual instructions, enabling joint multimodal learning. The LoRA configuration employed a rank of $r=16$, with a scaling parameter of $\alpha=16$ and a dropout probability of $0.05$ applied to the LoRA layers. LoRA adapters were applied to all linear layers of the model, while the original bias parameters were left unchanged. The task was configured for causal language modeling.

Finally, fine-tuning hyperparameters, such as learning rate, batch size, and number of epochs, to optimize results and prevent overfitting is also necessary. The fine-tuned model was deployed on a cloud-based platform, Google Colab \cite{parthasarathy2024ultimate}. Supervised fine-tuning was performed for five training epochs with a per-device batch size of 1. Gradient accumulation was conducted over four consecutive steps, resulting in an effective batch size of four samples per optimization step on each device. Model parameters were optimized using the fused PyTorch implementation of the AdamW optimizer with a learning rate of $2 \times 10^{-4}$. A constant learning-rate schedule was employed throughout training. To improve training stability and mitigate excessively large gradient updates, gradient clipping was applied with a maximum gradient norm of $0.3$.

\section{Results and Discussion}

\begin{table*}[t]

\centering
\caption{Performance comparison of Experiment 1.1 (Generalist) \& Experiment 1.2 (MoE) before and after fine-tuning. 
The $\pm k^\circ$ columns report accuracy within an angular tolerance of $k$ degrees.}
\label{tab1}

\resizebox{.9\textwidth}{!}{%
\begin{tabular}{lccccc|ccccc}
\hline
\multirow{2}{*}{\textbf{Model}} &
\multicolumn{5}{c|}{\textbf{Baseline}} &
\multicolumn{5}{c}{\textbf{Fine-tuned}} \\

\cmidrule(lr){2-6}
\cmidrule(lr){7-11}

& \textbf{Axis Acc.}
& \textbf{$\pm 0^\circ$}
& \textbf{$\pm 3^\circ$}
& \textbf{$\pm 5^\circ$}
& \textbf{$\pm 10^\circ$}
& \textbf{Axis Acc.}
& \textbf{$\pm 0^\circ$}
& \textbf{$\pm 3^\circ$}
& \textbf{$\pm 5^\circ$}
& \textbf{$\pm 10^\circ$} \\
\hline
Generalist
& 34.0\% & 0.0\% & 2.1\% & 2.1\% & 4.3\%
& 31.9\% & 0.0\% & 8.5\% & 17.0\% & 31.9\% \\
MoE
& 53.0\% & 6.3\% & 11.8\% & 12.5\% & 18.8\%
& 98.0\% & 18.8\% & 58.8\% & 87.5\% & 93.8\% \\
\hline
\end{tabular}%
}
\end{table*}

\textbf{(1) Results of 3D Rotation Process w/ Visual Cues: }In this experiment, the world coordinate system is incorporated in the images, providing spatial context for interpreting the arbitrary object’s rotation. AI inherently does not know what is considered the top or the bottom of an arbitrary object. VLMs are efficient multi-modal systems capable of jointly interpreting visual and textual information. However, their performance on purely visual tasks, such as object detection and rotation detection, may not match that of specialized computer vision models \cite{team2024gemini}. As a result, VLMs often rely heavily on textual guidance, especially for the coordinate system \cite{11050695}. The coordinate system makes communicating 3D rotation for VLM straightforward. As previously discussed, the JND for human angle discrimination is roughly 2° to 5° under ideal conditions. Therefore, we used thresholds ($\pm0^\circ$), ($\pm3^\circ$), ($\pm5^\circ$) and ($\pm10^\circ$) in this study (setting a lower and upper limit along with JND) to interpret as practical evaluation tolerances rather than using exactly human angular JND. Since recognizing rotations of complex three-dimensional objects is more demanding than discriminating simple lines or gratings, we evaluated AI's performance accuracy based on the thresholds. We used two approaches to fine-tuning here, and we can see a performance difference between the two approaches.

\textbf {Results of Experiment 1.1:} In the generalist approach, we fine-tuned the VLM for both axis and angle detection and tested with 47 images. Here, we notice that axis accuracy did not improve with fine-tuning; it slightly decreased (Table \ref{tab1}). Angle accuracy at ($\pm0^\circ$) tolerance remained 0\% before and after fine-tuning (Table 1). While considering ($\pm3^\circ$), ($\pm5^\circ$) and ($\pm10^\circ$) tolerance, we can notice some performance enhancement. However, a Chi-square test at p$<$0.05
indicated statistically significant improvement only at ($\pm5^\circ$) and ($\pm10^\circ$) tolerance between the generalist baseline model and the generalist fine-tuned model.
\par  $\chi^2$  (1, N = 94) = 0.8, \textit{p} $=$ .371 for ($\pm3^\circ$)\par
\par  $\chi^2$  (1, N = 94) = 4.0, \textit{p} $<$ .05 for ($\pm5^\circ$) \par
\par  $\chi^2$  (1, N = 94) = 8.47, \textit{p} $<$ .01 for ($\pm10^\circ$)\par

\par

These results indicate that the generalist fine-tuning method cannot train the VLM to achieve significant accuracy.

\par

\textbf {Results of Experiment 1.2:} From Table \ref{tab1}, we can see that the MoE fine-tuning method, where we separated axis detection and angle detection tasks, has drastically improved the VLM's performance. We used 47 images for testing purposes. Axis detection accuracy improved from 53.0\% to 98.0\% after fine-tuning. The fine-tuned MoE model achieved an axis-detection accuracy (98.0\%), which is more than three times that of the fine-tuned generalist model (31.9\%).
\par
In the case of angle detection, we can see a statistically significant difference for ($\pm3^\circ$), ($\pm5^\circ$) and ($\pm10^\circ$) at p$<$0.05 between the MoE baseline model and the MoE fine-tuned model.

\par  $\chi^2$  (1, N = 94) = 0.25, \textit{p} $=$ .617 for ($\pm0^\circ$)\par
\par  $\chi^2$  (1, N = 94) = 6.13, \textit{p} $<$ .05 for ($\pm3^\circ$)\par
\par  $\chi^2$  (1, N = 94) = 10.08, \textit{p} $<$ .01 for ($\pm5^\circ$) \par
\par  $\chi^2$  (1, N = 94) = 10.08, \textit{p} $<$ .01 for ($\pm10^\circ$)

\textbf{Generalist vs. MoE Fine-tuning}: If we consider performance improvement between the generalist fine-tuned model and the MoE fine-tuned model, we see a statistically significant performance difference across all criteria (Table \ref{tab1}).

\par  $\chi^2$  (1, N = 94) = 29.03, \textit{p} $<$ .0001 for axis detection \par 
\par  $\chi^2$  (1, N = 94) = 7.11, \textit{p} $<$ .01 for ($\pm0^\circ$)\par
\par  $\chi^2$  (1, N = 94) = 20.83, \textit{p} $<$ .001 for ($\pm3^\circ$)\par
\par  $\chi^2$  (1, N = 94) = 27.68, \textit{p} $<$ .001 for ($\pm5^\circ$) \par
\par  $\chi^2$  (1, N = 94) = 25.68, \textit{p} $<$ .001 for ($\pm10^\circ$)\par

These results indicate that the MoE architecture better capture distinct visual and spatial patterns by routing inputs through specialized expert components, as model size, training data, and fine-tuning conditions were similar (Table \ref{tab1}). Even though we describe rotation by mentioning the angle and axis of rotation together, training AI separately for axis and angle demonstrated better performance.

\textbf{(2) Results of 2D Rotation Process w/o Visual Cues}: As this dataset contains 2D planar view of 3D objects, the rotation is conducted along one axis, and there is no coordinate system displayed for the vision model. Therefore, we are only considering VML's angle detection capacity for this part. This dataset also consists of identifiable objects along with arbitrary objects. The angles in this dataset increment by 5°. In this test, we investigate whether fine-tuning can enable a VLM to learn rotation angles without relying on an explicit coordinate system. We are also investigating whether a fine-tuned VLM demonstrates better performance for identifiable objects than arbitrary objects. \par

If we observe the mean results from Table \ref{tab2}, the fine-tuned model improved considerably compared to the baseline for both identifiable and arbitrary objects. For the fine-tuned model of arbitrary objects, $\pm0^\circ$, $\pm5^\circ$, and $\pm10^\circ$ do not have any significant difference in accuracy. This implies that most correct detection cases were very close (less than $\pm5^\circ$) to the true answers.

At $\pm5^\circ$ and $\pm10^\circ$ tolerance levels, fine-tuned identifiable and arbitrary objects do not show significant performance differences (Table \ref{tab2}). However, at $\pm0^\circ$ level, we can observe performance differences between identifiable and arbitrary objects, where identifiable objects underperform. This implies the pretrained model's capability to identify the object does not affect angle detection. Arbitrary objects with more linear features enhance the performance.\par

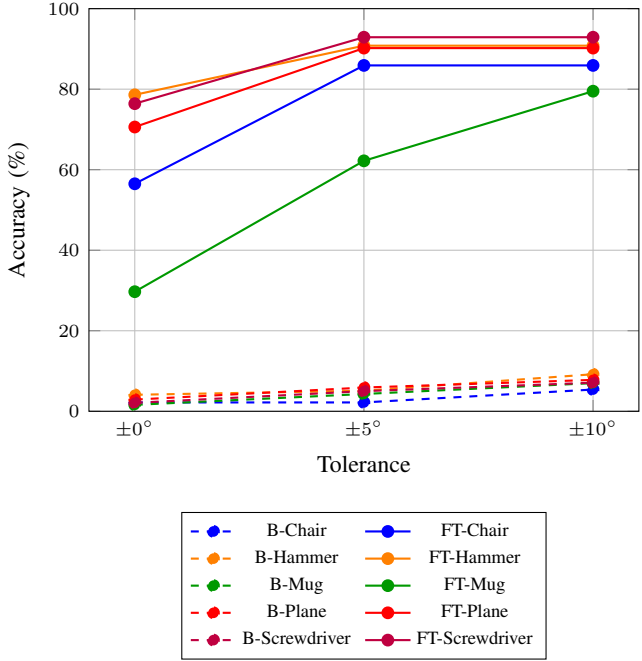
\begin{figure}[t]
\centering

\begin{tikzpicture}
\begin{axis}[
    width=\columnwidth,
    height=0.78\columnwidth,
    xlabel={Tolerance},
    ylabel={Accuracy (\%)},
    symbolic x coords={pm0,pm5,pm10},
    xtick=data,
    xticklabels={
        $\pm0^\circ$,
        $\pm5^\circ$,
        $\pm10^\circ$
    },
    ymin=0,
    ymax=100,
    ytick={0,20,40,60,80,100},
    grid=major,
    tick label style={font=\scriptsize},
    label style={font=\small},
    legend style={
        font=\scriptsize,
        at={(0.5,-0.25)},
        anchor=north,
        legend columns=2,
        /tikz/every even column/.append style={
            column sep=3pt
        }
    }
]

\addplot[
    blue,
    dashed,
    thick,
    mark=*
] coordinates {
    (pm0,2.20)
    (pm5,2.20)
    (pm10,5.40)
};
\addlegendentry{B-Chair}

\addplot[
    blue,
    solid,
    thick,
    mark=*
] coordinates {
    (pm0,56.50)
    (pm5,85.90)
    (pm10,85.90)
};
\addlegendentry{FT-Chair}

\addplot[
    orange,
    dashed,
    thick,
    mark=*
] coordinates {
    (pm0,4.10)
    (pm5,5.10)
    (pm10,9.20)
};
\addlegendentry{B-Hammer}

\addplot[
    orange,
    solid,
    thick,
    mark=*
] coordinates {
    (pm0,78.60)
    (pm5,90.80)
    (pm10,90.80)
};
\addlegendentry{FT-Hammer}

\addplot[
    green!60!black,
    dashed,
    thick,
    mark=*
] coordinates {
    (pm0,1.60)
    (pm5,4.30)
    (pm10,7.00)
};
\addlegendentry{B-Mug}

\addplot[
    green!60!black,
    solid,
    thick,
    mark=*
] coordinates {
    (pm0,29.70)
    (pm5,62.20)
    (pm10,79.50)
};
\addlegendentry{FT-Mug}

\addplot[
    red,
    dashed,
    thick,
    mark=*
] coordinates {
    (pm0,2.90)
    (pm5,5.90)
    (pm10,7.80)
};
\addlegendentry{B-Plane}

\addplot[
    red,
    solid,
    thick,
    mark=*
] coordinates {
    (pm0,70.60)
    (pm5,90.20)
    (pm10,90.20)
};
\addlegendentry{FT-Plane}

\addplot[
    purple,
    dashed,
    thick,
    mark=*
] coordinates {
    (pm0,2.10)
    (pm5,5.00)
    (pm10,7.10)
};
\addlegendentry{B-Screwdriver}

\addplot[
    purple,
    solid,
    thick,
    mark=*
] coordinates {
    (pm0,76.40)
    (pm5,92.90)
    (pm10,92.90)
};
\addlegendentry{FT-Screwdriver}

\end{axis}
\end{tikzpicture}

\caption{Experiment 2.1 Results: Baseline and fine-tuned accuracy for five identifiable objects under angular tolerances of $\pm0^\circ$, $\pm5^\circ$, and $\pm10^\circ$. Dashed lines represent baseline (B) performance, while solid lines represent fine-tuned (FT) performance.}
\label{fig:Known obj res}
\end{figure}

\textbf {Results of Experiment 2.1:} Fig. \ref{fig:Known obj res} demonstrates a substantial improvement in model performance after fine-tuning for each object, even for $\pm0^\circ$ tolerance. There are not many performance differences between  $\pm5^\circ$ and $\pm10^\circ$ tolerance levels for the fine-tuned model for most objects (lines between $\pm5^\circ$ and $\pm10^\circ$ are almost horizontal). Only the mug in Fig. \ref{fig:Known obj res} shows improvement between $\pm5^\circ$ and $\pm10^\circ$ tolerance levels. The reason for that could be that the mug's top view has fewer linear features than other objects for angle detection, which makes the detection more difficult, but the fine-tuning has achieved more gains in accuracy in looser tolerance cases.

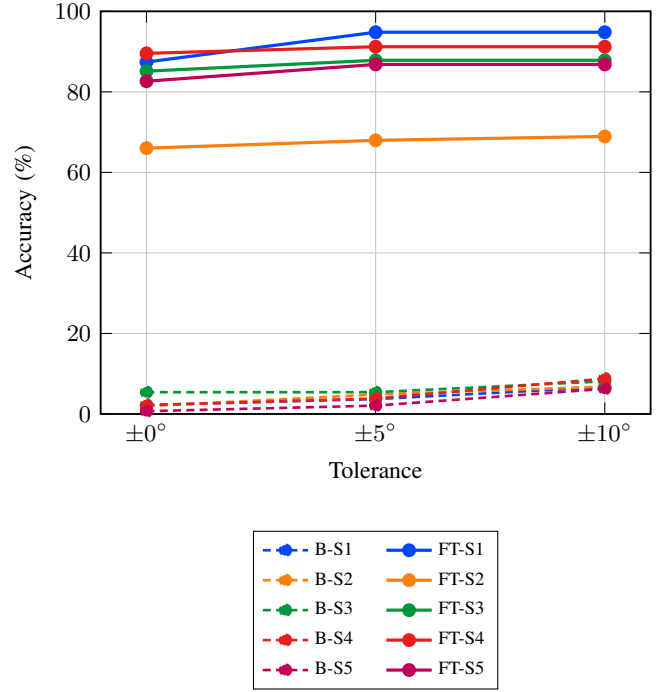
\begin{figure}[t]
\centering

\begin{tikzpicture}
\begin{axis}[
    width=\columnwidth,
    height=0.78\columnwidth,
    ymin=0,
    ymax=100,
    ytick={0,20,40,60,80,100},
    ylabel={Accuracy (\%)},
    xlabel={Tolerance},
    symbolic x coords={pm0,pm5,pm10},
    xtick=data,
    xticklabels={
        $\pm0^\circ$,
        $\pm5^\circ$,
        $\pm10^\circ$
    },
    grid=major,
    grid style={
        draw=gray!45,
        line width=0.4pt
    },
    axis line style={
        draw=black,
        line width=0.7pt
    },
    tick align=inside,
    tick style={black},
    label style={font=\small},
    tick label style={font=\small},
    legend style={
        at={(0.5,-0.29)},
        anchor=north,
        legend columns=2,
        draw=black,
        fill=white,
        font=\scriptsize,
        row sep=1pt,
        /tikz/every even column/.append style={
            column sep=0.35cm
        }
    }
]

\addplot[
    color=shapeone,
    densely dashed,
    line width=1pt,
    mark=*,
    mark size=2pt
] coordinates {
    (pm0,2.22)
    (pm5,3.70)
    (pm10,6.67)
};
\addlegendentry{B-S1}

\addplot[
    color=shapeone,
    solid,
    line width=1.2pt,
    mark=*,
    mark size=2pt
] coordinates {
    (pm0,87.41)
    (pm5,94.81)
    (pm10,94.81)
};
\addlegendentry{FT-S1}

\addplot[
    color=shapetwo,
    densely dashed,
    line width=1pt,
    mark=*,
    mark size=2pt
] coordinates {
    (pm0,1.94)
    (pm5,4.85)
    (pm10,6.80)
};
\addlegendentry{B-S2}

\addplot[
    color=shapetwo,
    solid,
    line width=1.2pt,
    mark=*,
    mark size=2pt
] coordinates {
    (pm0,66.02)
    (pm5,67.96)
    (pm10,68.93)
};
\addlegendentry{FT-S2}

\addplot[
    color=shapethree,
    densely dashed,
    line width=1pt,
    mark=*,
    mark size=2pt
] coordinates {
    (pm0,5.41)
    (pm5,5.41)
    (pm10,8.11)
};
\addlegendentry{B-S3}

\addplot[
    color=shapethree,
    solid,
    line width=1.2pt,
    mark=*,
    mark size=2pt
] coordinates {
    (pm0,85.14)
    (pm5,87.84)
    (pm10,87.84)
};
\addlegendentry{FT-S3}

\addplot[
    color=shapefour,
    densely dashed,
    line width=1pt,
    mark=*,
    mark size=2pt
] coordinates {
    (pm0,2.20)
    (pm5,3.85)
    (pm10,8.79)
};
\addlegendentry{B-S4}

\addplot[
    color=shapefour,
    solid,
    line width=1.2pt,
    mark=*,
    mark size=2pt
] coordinates {
    (pm0,89.56)
    (pm5,91.21)
    (pm10,91.21)
};
\addlegendentry{FT-S4}

\addplot[
    color=shapefive,
    densely dashed,
    line width=1pt,
    mark=*,
    mark size=2pt
] coordinates {
    (pm0,0.69)
    (pm5,2.08)
    (pm10,6.25)
};
\addlegendentry{B-S5}

\addplot[
    color=shapefive,
    solid,
    line width=1.2pt,
    mark=*,
    mark size=2pt
] coordinates {
    (pm0,82.64)
    (pm5,86.81)
    (pm10,86.81)
};
\addlegendentry{FT-S5}

\end{axis}
\end{tikzpicture}

\caption{Experiment 2.2 Results: Baseline and fine-tuned accuracy for five arbitrary objects under angular tolerances of $\pm0^\circ$, $\pm5^\circ$, and $\pm10^\circ$. Dashed lines represent baseline (B) performance, while solid lines represent fine-tuned (FT) performance.}
\label{fig:Unknown obj res}
\end{figure}

\textbf {Results of Experiment 2.2:} Fig. \ref{fig:Unknown obj res} demonstrates a substantial improvement in model performance after fine-tuning for each arbitrary object, even for $\pm0^\circ$ tolerance. There are not many performance differences between  $\pm5^\circ$ and $\pm10^\circ$ tolerance levels for the fine-tuned model for most objects (lines between $\pm5^\circ$ and $\pm10^\circ$ are almost horizontal).

\begin{table}[t]
\centering
\caption{Mean performance comparison of Experiment 2.1 (Identifiable objects) \& Experiment 2.2 (Arbitrary objects) before and after fine-tuning.}
\label{tab2}

\resizebox{\columnwidth}{!}{%
\begin{tabular}{lccc|ccc}
\toprule

\multirow{2}{*}{\textbf{Object}} &
\multicolumn{3}{c|}{\textbf{Baseline}} &
\multicolumn{3}{c}{\textbf{Fine-tuned Model}} \\

\cmidrule(lr){2-4}
\cmidrule(lr){5-7}

& \textbf{$\pm0^\circ$}
& \textbf{$\pm5^\circ$}
& \textbf{$\pm10^\circ$}
& \textbf{$\pm0^\circ$}
& \textbf{$\pm5^\circ$}
& \textbf{$\pm10^\circ$} \\

\midrule

Identifiable
(Exp 2.1)
& 2.58\%
& 4.50\%
& 7.30\%
& 62.36\%
& 84.40\%
& 87.86\% \\

Arbitrary (Exp 2.2)
& 2.49\%
& 3.98\%
& 7.32\%
& 82.15\%
& 85.73\%
& 85.92\% \\

\bottomrule
\end{tabular}%
}
\end{table}

\section{Conclusions and Future Work}

Spatial skills are critically important in STEM and art (STEAM) fields like architecture, computer graphics, civil engineering, and medicine, etc. 
This paper investigates a fine-tuning method to enhance VLM's proficiency in understanding 2D \& 3D rotations. We generated multiple object rotation datasets for training and testing purposes. For detecting 2D \& 3D rotations, the experiments showed very promising results, where the fine-tuned model performed significantly better than the base model. In Experiments 1.1 \& 1.2, the MoE fine-tuning methods showed significantly better performance than generalist fine-tuning for rotation detection, even though we describe rotation by specifying both the angle and axis of rotation. In Experiments 2.1 and 2.2, we also found that for 2D representation of 3D objects, the fine-tuning method improves the VLM's angle detection capability significantly, and the improvements apply to both identifiable and arbitrary objects similarly, indicating that the proposed AI fine-tuning method is effective for detecting rotation angles of both known and unknown objects. The improvements are more significant for difficult test cases such as objects without many linear features.
\par

This work can be further improved by including testing with larger datasets, the latest AI models, etc. AI’s spatial ability can be further enhanced by training models on images depicting transformations such as rotation, translation, scaling, and changes in perspective. Diverse image or video representations of arbitrary physical 3D objects, with or without contextual information, may further strengthen these capabilities.\par

AI with spatial capability can be beneficial in the future in education, training, and practice in STEAM and the AEC industry. For instance, in construction and fabrication settings, generative AI could monitor workers’ activities in real time and deliver step-by-step assembly guidance. It can use its vision capability to detect errors in construction sites and provide solutions. Similarly, generative AI could support students by providing immediate feedback on rotation angles and object dimensions, thereby facilitating more effective learning of spatial transformations, as well as their mathematical representations. In addition, improved AI's spatial intelligence is expected to enhance immersive AR/VR-assisted manufacturing.

\section*{Acknowledgment}

Omitted for blind review.

\bibliographystyle{IEEEtran}
\bibliography{refs}

\end{document}